\documentclass{article}
\usepackage{spconf,amsmath,epsfig}
\usepackage{float}
\usepackage{hyperref}
\usepackage{geometry} 
\usepackage{lscape}
\usepackage{amssymb}
\usepackage{caption}
\usepackage{afterpage} 
\usepackage{colortbl} 
\usepackage{xcolor}   
\usepackage{multirow}
\usepackage{geometry}  
\usepackage{graphicx}
\usepackage{lipsum}
\usepackage{tabularx}

\usepackage{booktabs}
\usepackage{siunitx}
\usepackage{microtype}

\DeclareMathOperator*{\argmin}{arg\,min}
\title{Multi-Label Scene Classification in Remote Sensing Benefits from Image Super-Resolution}
\name{Ashitha Mudraje$^{1,2}$ \hspace{0.6em} Brian B. Moser$^{1,2}$ \hspace{0.6em} Stanislav Frolov$^{1,2}$ \hspace{0.6em} Andreas Dengel$^{1,2}$}
\address{$^1$German Research Center for Artificial Intelligence, Germany\\
$^2$RPTU Kaiserslautern-Landau, Germany \\
{\tt\small first.last@dfki.de}
}

\begin{document}
%
\definecolor{highlightyellow}{rgb}{1.0, 1.0, 0.6} 
\definecolor{highlightgreen}{rgb}{0.5, 1.0, 0.5}  
\definecolor{highlightcyan}{rgb}{0.5, 1.0, 0.87}  
\maketitle
\begin{abstract}
Satellite imagery is a cornerstone for numerous Remote Sensing (RS) applications; however, limited spatial resolution frequently hinders the precision of such systems, especially in multi-label scene classification tasks as it requires a higher level of detail and feature differentiation.
In this study, we explore the efficacy of image Super-Resolution (SR) as a pre-processing step to enhance the quality of satellite images and thus improve downstream classification performance.
We investigate four SR models - SRResNet, HAT, SeeSR, and RealESRGAN - and evaluate their impact on multi-label scene classification across various CNN architectures, including ResNet-50, ResNet-101, ResNet-152, and Inception-v4.
Our results show that applying SR significantly improves downstream classification performance across various metrics, demonstrating its ability to preserve spatial details critical for multi-label tasks.
Overall, this work offers valuable insights into the selection of SR techniques for multi-label prediction in remote sensing and presents an easy-to-integrate framework to improve existing RS systems.
\end{abstract}
\section{Introduction}
\label{sec:intro}
Remote Sensing (RS) is vital for monitoring and analyzing the Earth’s surface \cite{meng2023single}. 
However, despite the increasing demand for High-Resolution (HR) imagery in this domain, limited sensor capabilities often constrain fine-grained classification, detection, or mapping tasks \cite{wang2023remote,dong2022real}.
To alleviate these constraints - without relying solely on costly HR sensors or upgrades - image Super-Resolution (SR) offers a cost-efficient alternative by generating HR images from Lower-Resolution (LR) inputs \cite{salvetti2020multi}.
While SR has been explored for various classification tasks \cite{shah2024webcam,chen2024super,he2024connecting}, its influence on multi-label scene prediction in RS remains underexplored.

\begin{figure}[t!]
	\centering
	\includegraphics[width=.9\linewidth]{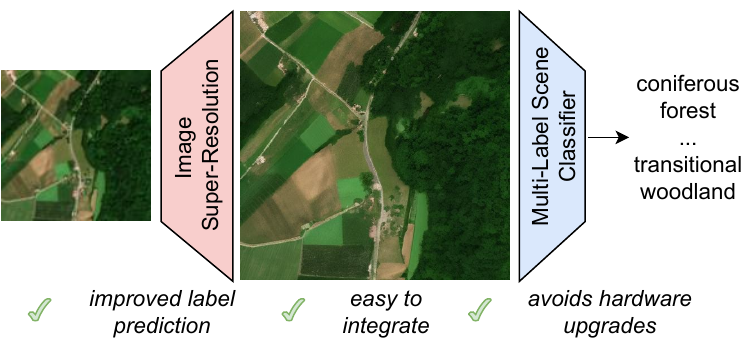}
	\caption{Illustration of our proposed pipeline that uses image super-resolution as a pre-processing step for multi-label scene classification for improved label prediction.}\label{fig:idea}
\end{figure}

In RS, multi-label scene classification typically involves detecting multiple land cover types, urban structures, vegetation, or water bodies within a single satellite image, necessitating the capture of fine spatial details across diverse scales \cite{cheng2024mmdl,sumbul2020deep}. 
Yet, with LR images, classical approaches often fail to capture subtle boundaries or small objects, ultimately leading to diminished precision \cite{dong2022real}. 
By applying SR as pre-processing step, shown in \autoref{fig:idea}, models can leverage these finer features, leading to improved multi-label recognition \cite{shah2024webcam}.

This paper investigates the potential of SR to enhance multi-label scene classification in RS. 
We selected a range of pre-trained SR techniques - including diffusion-based, GAN-based, and CNN approaches - and compare their performance on SR-enhanced and original LR images \cite{ledig2017photo,chen2023hat,wu2024seesr,wang2021realesrgan}. 
Our findings highlight that using image SR excels at reconstructing spatial details critical for classification. 
As a result, the added details lead to improved identification of multiple labels within a single image. 
In summary, we shed light on the benefits and limitations of SR in improving downstream prediction accuracy.

\section{Related work}
\label{sec:format}

\subsection{Remote Sensing}
Singh et al. \cite{singh2022multilabelclassificationremotesensingimages} explore the application of satellite imagery for monitoring environmental changes. The authors leverage various machine learning and deep learning models such as ResNet\cite{he2016identity}, VGG\cite{simonyan2014very}, Inception\cite{szegedy2015going}, and so on to classify satellite images based on atmospheric conditions and land use. By employing multi-label classification, the authors captured the intricate relationships between different environmental factors. 

Similar in spirit, Liu et al. \cite{liu2021multi} presents a simplified residual network model for multi-label classification of RS images. The model leverages transfer learning with ResNet-50, incorporating some techniques like  batch  normalization, image augmentation and self-defined loss evaluation index to enhance training efficiency and accuracy. The proposed approach effectively addresses challenges related to power consumption and performance, achieving over 90\% precision and recall

Similarly, Gardner et al. \cite{gardner2017multi} explored multi-label classification by applying CNN-based architectures like VGG-16, Inception-v3, and ResNet-50 to classify Amazon rainforest features and illegal mining activities. Although a baseline model captured common categories, it struggled with rare classes crucial for detecting illicit operations. Notably, ResNet-50 outperformed other architectures, demonstrating the value of deeper networks for multi-label remote sensing tasks.

While deeper CNN architectures have yielded promising results, the question of how SR might further boost downstream classification performance in RS remains underexplored. 
The present work addresses this gap by systematically investigating the impact of SR-driven enhancements.

\subsection{Image Super-Resolution}
A trained SR model $M_\theta: \mathbb{R}^{\text{H} \times \text{W} \times \text{C}} \rightarrow \mathbb{R}^{s\cdot\text{H} \times s\cdot\text{W} \times \text{C}}$ should inverse the degradation relationship between a LR image $\mathbf{x} \in \mathbb{R}^{\text{H} \times \text{W} \times \text{C}}$ and the HR image $\mathbf{y} \in \mathbb{R}^{s\cdot\text{H} \times s\cdot\text{W} \times \text{C}}$, where $s$ denotes the scaling. 
The optimization of $\theta$ is based on a dataset $\mathbb{D}_{SR} = \{ \left( \mathbf{x}_i, \mathbf{y}_i\right)\}^N_{i=1}$ of $N$ LR-HR pairs with the goal
\begin{equation}
    \theta^{*} = \argmin_\theta \mathbb{E}_{(\mathbf{x}_i, \mathbf{y}_i) \in \mathbb{D}_{SR}}\lVert M_\theta (\mathbf{x}_i) - \mathbf{y}_i\rVert^2.
\end{equation}

Using SR have found use across diverse domains, ranging from medical imaging, where sharper images can critically affect patient outcomes, to satellite imagery, enabling more precise geographic analysis of the Earth’s surface \cite{song2022deep, tang2021single,meng2023single}.
This work investigates the influence of using existing and pre-trained SR methods as a pre-processing stage to RS downstream tasks.

\section{Methodology}
Our goal is to apply SR models of the form $M_\theta: \mathbb{R}^{\text{H} \times \text{W} \times \text{C}} \rightarrow \mathbb{R}^{s\cdot\text{H} \times s\cdot\text{W} \times \text{C}}$ prior to training a multi-label classifier to improve the precision of RS downstream tasks.
Since the amount and quality of generally available images outnumber high-quality satellite images, we refer to pre-trained SR models with fixed parameters $\theta$ rather than training a new model $M_\theta$ from scratch.

\subsection{SR Models}
\label{sec:majhead}
For SR methods, two primary factors drive performance: the model architecture $M_\theta$ and the training objectives to optimize $\theta$ \cite{moser2023hitchhiker}.
For the latter, SR models can be categorized into two groups: regression-based models, which typically employ a regression loss, and generative SR models (GANs and diffusion models) \cite{moser2024diffusion}.

Consequently, we analyze representatives of each category.
For regression-based models, we employ the ResNet-based model SRResNet \cite{ledig2017photo,he2016deep} and the vision transformer HAT \cite{chen2023activating,chen2023hat}. 
For generative SR, we analyze SeeSR \cite{wu2024seesr} as a diffusion-based representative and RealESRGAN \cite{wang2021realesrgan} as a representative for GANs.
As image SR models are usually trained for 2$\times$, 3$\times$, or 4$\times$, we will use 4$\times$ pre-trained models to allow for maximum flexibility for the multi-label scene classifier.

\subsection{Multi-Label Classifier}
We adopt four commonly used models for multi-label scene classification, namely ResNet-50, ResNet-101, ResNet-152, and Inception-v4. 
We train each model under two configurations:
\begin{itemize}
    \item \textbf{Baseline (No SR):} The network is trained directly on the original LR images ($120\times120$).
    \item \textbf{With SR Pre-processing (SR):} The network is trained on images super-resolved by one of the four SR models described in the previous Section (i.e., SRResNet, HAT, SeeSR, or RealESRGAN). We first apply the respective SR model ($4\times$ i.e. $480\times480$ resolution) and then feed the enhanced images to the multi-label classifier.
\end{itemize}


\subsection{Impact Assessment}
To evaluate the impact of image SR models on the classifier under different aspects, we employ the following evaluation metrics: 
\begin{itemize}
    \item \textbf{Sample Accuracy (ACC):} Measures the proportion of correctly predicted labels among all labels. Giving equal importance to each sample of the test set.
    \item \textbf{Hamming Loss (HL):} Quantifies the fraction of misclassified labels, capturing the multi-label misalignment \cite{mao2024multilabellearningstrongerconsistency}. Lower HL indicates fewer label-wise errors.
    \item \textbf{One-Error (OE):} Checks whether the top prediction (the label with the highest probability) is present in the true label set. A lower OE implies the model’s highest-confidence prediction is more likely correct.
    \item \textbf{Precision (P)}, \textbf{Recall (R)}, and \textbf{F1-Score:} Standard measures assessing the balance between correctly predicted labels (Precision) and the coverage of positive instances (Recall). F1 is their harmonic mean. All metrics are calculated based on the sample (test) data.
    \item \textbf{Macro F2 Score:} An extension of F1 that places additional emphasis on Recall. Useful when missing labels is costlier than having false positives.
\end{itemize}
\noindent

\begin{table*}[t!]
	\centering
	\caption{Accuracy (ACC), Hamming Loss (HL), One-Error (OE), Precision (P), Recall (R), and F1 Score Comparison Across Models and Configurations. Best (classifier-wise) results are marked in bold.}
    \label{tab:main}
    \resizebox{\linewidth}{!}{%
	\begin{tabular}{lcccccc|cccccc|cccccc|cccccc}
        \multirow{2}{*}{} & \multicolumn{6}{c|}{\textbf{ResNet-50}} & \multicolumn{6}{c|}{\textbf{ResNet-101}} & \multicolumn{6}{c|}{\textbf{ResNet-152}} & \multicolumn{6}{c}{\textbf{Inception-v4}} \\
         & \textbf{ACC} & \textbf{HL} & \textbf{OE} & \textbf{P} & \textbf{R} & \textbf{F1} & \textbf{ACC} & \textbf{HL} & \textbf{OE} &\textbf{P} & \textbf{R} & \textbf{F1} & \textbf{ACC} & \textbf{HL} & \textbf{OE} &\textbf{P} & \textbf{R} & \textbf{F1} & \textbf{ACC} & \textbf{HL} & \textbf{OE} &\textbf{P} & \textbf{R} & \textbf{F1} \\
        \midrule
        No SR  & 0.411 & 0.138 & 0.323 & 0.556 & 0.564 & 0.526 & 0.413 & 0.144 & 0.345 & 0.534 & 0.570 & 0.525 & 0.402 & 0.137 & 0.371 & 0.533 & 0.546 & 0.512 & 0.278 & 0.153 & 0.498 & 0.458 & 0.327 & 0.360 \\
        SRResNet  & \textbf{0.474} & 0.138 & \textbf{0.231} & \textbf{0.647} & \textbf{0.586} & \textbf{0.583} & \textbf{0.462} & 0.126 & \textbf{0.272} & 0.609 & \textbf{0.592} & \textbf{0.571} & 0.419 & 0.144 & 0.295 & 0.550 & \textbf{0.587} & 0.535 & 0.333 & \textbf{0.136} & \textbf{0.330} & \textbf{0.547} & 0.381 & 0.422 \\
        HAT      & 0.448 & 0.125 & 0.265 & 0.617 & 0.570 & 0.559 & 0.457 & \textbf{0.119} & 0.278 &\textbf{0.619} & 0.569 & 0.564 & \textbf{0.464} & \textbf{0.115} & \textbf{0.241} & \textbf{0.648} & 0.560 & \textbf{0.570} & \textbf{0.421} & 0.159 & 0.338 & 0.530 & \textbf{0.585} & \textbf{0.531} \\

        SeeSR & 0.445 & \textbf{0.122} & 0.290 & 0.606 & 0.561 & 0.552 & 0.446 & 0.122 & 0.291 & \textbf{0.613} & 0.555 & \textbf{0.572} & 0.452 & 0.122 & 0.295 & 0.595 & \textbf{0.587} & 0.562 & 0.233 & 0.171 & 0.540 & 0.361 & 0.295 & 0.306 \\
        RealESRGAN & 0.439 & 0.124 & 0.277 & 0.622 & 0.555 & 0.550 & 0.440 & 0.122 & 0.284 & 0.598 & 0.566 & 0.550 & 0.455 & 0.120 & 0.265 & 0.626 & 0.561 & 0.560 & 0.325 & 0.142 & 0.437 & 0.487 & 0.412 & 0.419 \\
        
	\end{tabular}
    }
\end{table*}

\section{Experiments}


We utilized a standard dataset for multi-label prediction in remote sensing for our experiments: BigEarthNet \cite{sumbul2019bigearthnet,sumbul2021bigearthnet}, containing 519,284 non-overlapping image patches, where CORINE Land Cover (CLC)\cite{buttner2004corine} database provides one or more land cover class labels (multi-labels) for each image\cite{sumbul2020bigearthnet}. Each patch is a segment of 120 × 120 pixels for bands of 10m. These 10m band patches stacked to make RGB images(LR images).



\subsection{Quantitative Results}

\autoref{tab:main} shows the quantitative results.
In short, training on SR-enhanced images outperforms the baseline across all classifier backbones, confirming the value of SR in recovering details beneficial for multi-label prediction. 
Notably, SRResNet achieves the highest accuracy on ResNet-50 and ResNet-101, demonstrating its strong performance in moderately deep networks. 
Meanwhile, HAT attains the best Hamming Loss on ResNet-101 and ResNet-152, indicating more precise label-wise predictions when paired with deeper architectures.

\begin{table}[t!]
	\centering
	\caption{ Macro F2 Score for 19 classes, evaluated on ResNet-152 model. Best (label-wise) is marked in bold. }
    \label{tab:macro_f2}
    \resizebox{\linewidth}{!}{%
	\begin{tabular}{l c|c|c|c|c}
        \textbf{Class} & \textbf{No SR} & \textbf{SRResNet} & \textbf{HAT} & \textbf{SeeSR} & \textbf{RealESRGAN} \\
        \midrule
        Urban fabric                                      & 0.519  & \textbf{0.534} & 0.428  & 0.506  & 0.452  \\ 
        Industrial or commercial units                   & 0.249  & 0.235          & 0.256  & 0.179  & \textbf{0.258}  \\ 
        Arable land                                      & 0.701  & 0.732 & \textbf{0.765}  & 0.756 & 0.730  \\ 
        Permanent crops                                  & 0.356  & 0.374 & 0.341  & \textbf{0.408}  & 0.257  \\ 
        Pastures                                         & 0.374  & 0.389 & 0.365  & \textbf{0.448}  & 0.416  \\ 
        Complex cultivation patterns                     & \textbf{0.633}  & 0.512          & 0.556  & 0.628  & 0.488  \\ 
        Land principally occupied by agriculture         & \textbf{0.556}  & 0.537          & 0.383  & 0.446  & 0.168  \\ 
        Agro-forestry areas                              & 0.588  & 0.637          & \textbf{0.716} & 0.668  & 0.643  \\ 
        Broad-leaved forest                              & 0.555  & 0.535          & 0.460  & 0.554  & \textbf{0.577}  \\ 
        Coniferous forest                                & 0.653  & 0.618          & 0.725  & 0.720  & \textbf{0.742}  \\ 
        Mixed forest                                     & \textbf{0.754}  & 0.552          & 0.568  & 0.719  & 0.729  \\ 
        Natural grassland \& sparsely vegetated areas   & 0.013  & 0.037          & 0.013  & 0.0004  & \textbf{0.102}  \\ 
        Moors, heathland \& sclerophyllous vegetation   & 0.178  & \textbf{0.285} & 0.168  & 0.044  & 0.084  \\ 
        Transitional woodland, shrub                     & 0.520  & \textbf{0.599} & 0.521  & 0.478  & 0.487  \\ 
        Beaches, dunes, sands                            & 0.274  & 0.276          & \textbf{0.449}  & 0.366  & 0.324  \\ 
        Inland wetlands                                  & 0.244  & 0.176          & 0.157  & 0.221  & \textbf{0.261}  \\ 
        Coastal wetlands                                 & 0.082  & 0.111          & \textbf{0.226}  & 0.045  & 0.089  \\ 
        Inland waters                                    & 0.575  & 0.662 & 0.678  & \textbf{0.683}  & 0.642  \\ 
        Marine waters                                    & 0.089  & \textbf{0.602} & 0.597  & 0.430  & 0.535  \\ \midrule
        \textbf{Average Macro F2 score}            & 0.417  & \textbf{0.443} & 0.440  & 0.437  & 0.420  \\ 
	\end{tabular}
    }
\end{table}

Regarding One-Error, SRResNet provides the largest reduction on ResNet-50 (0.231 vs.\ 0.323 baseline). However, with deeper models such as ResNet-152, HAT outperforms SRResNet (0.241 vs.\ 0.295). This finding suggests that while SRResNet excels in shallower configurations, HAT’s attention mechanisms align better with higher-capacity networks.
%
%
%
Similarly,  SRResNet consistently yields high F1 scores on ResNet-50, partly due to a robust balance of Precision and Recall. 
In contrast, HAT demonstrates stronger Precision and F1 in deeper setups (ResNet-152, Inception-v4). 

\begin{figure}[h!]
    \centering
    \includegraphics[width=\columnwidth]{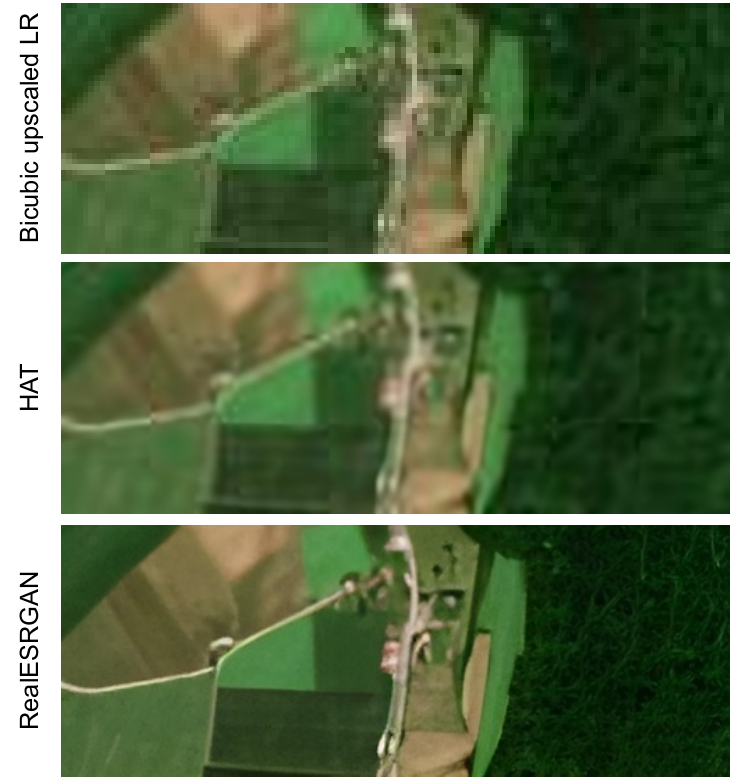}
    \caption{SR Comparison of HAT and RealESRGAN. While HAT provides relatively balanced enhancements, RealESRGAN tends to hallucinate details (e.g., overemphasizing streets and introducing artificial patterns in forest regions), illustrating the pitfalls of generative SR methods in certain remote sensing scenes.}
    \label{fig:example}
\end{figure}

Overall, these three perspectives (ACC/HL, OE, and P/R/F1) show consistent performance trends yet emphasize different quality aspects. 
Despite their complementary perspectives, they collectively indicate that attention-based SR (i.e., HAT) delivers the strongest gains when paired with deeper networks. 
In contrast, SRResNet provides the best quality for smaller architectures.
While all tested SR methods generally enhance multi-label prediction, generative approaches (i.e., SeeSR and RealESRGAN) hallucinate details, which explains their reduced positive impact, as exemplified in \autoref{fig:example}.

By examining the class-level results via Macro F2 (see \autoref{tab:macro_f2}), we observe that SR notably boosts performance for certain land-cover types, particularly those defined by clear boundaries and texture (e.g., Marine and Inland Waters). 
Yet, while a SR method improves certain classes, it does not consistently improve predictions across all labels. 
These variations suggest that SR’s effectiveness can be class-specific and should be factored into pre-processing decisions for multi-label RS tasks.





\begin{figure*}[t!]
    \centering
    \includegraphics[width=\textwidth]{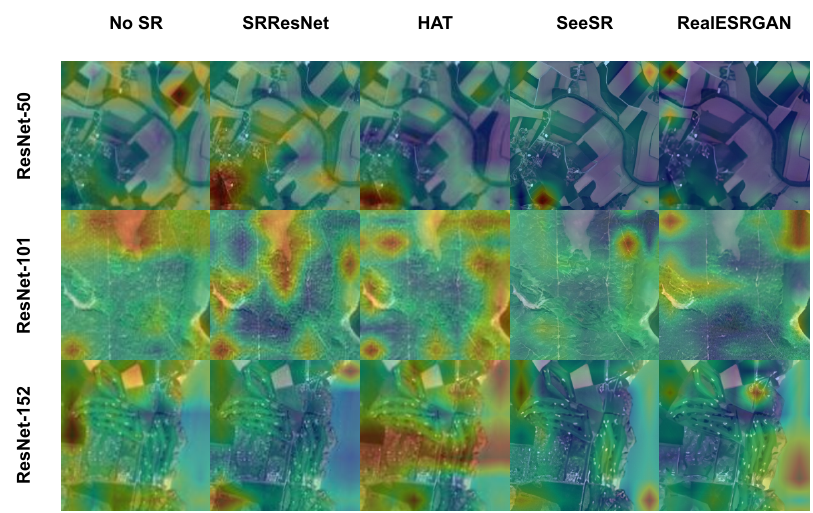}
\caption{Comparison of Grad-CAM visualizations in ResNet models between the baseline (No SR) and various SR methods. SRResNet (for shallow classifier) and HAT (for deeper classifier) lead to more activation coverage across the whole satellite image.}
    \label{fig:cam}
\end{figure*}

\subsection{Qualitative Results}

To gain deeper insight into how SR pre-processing influences network behavior, we employ Grad-CAM \cite{selvaraju2017grad} to visualize class activation maps in the final convolutional layer of ResNet architectures. 
The results are shown in \autoref{fig:cam}.
Overall, SR-enhanced images exhibit more pronounced and varied activations (highlighted by dark red or blue regions), indicating that the classifier focuses more strongly on distinct features.
Interestingly, the strongest performance gains often coincide with broadly distributed positive CAM responses: for ResNet-50, SRResNet yields widespread high-intensity activations, whereas for ResNet-152, HAT demonstrates similarly extensive coverage.
These observations align with the quantitative results, suggesting that spatially richer activations under SR pre-processing directly contribute to improved multi-label classification.

One plausible explanation for the observed performance gains and heightened activation variance is that higher resolution inputs better align with the receptive field that increases with deeper networks, allowing the classifier to extract richer, more discriminative features at each convolutional layer. 
Consequently, the classifier can more effectively utilize these cues when predicting multiple labels, leading to a broader and more varied activation map in the Grad-CAM visualizations and ultimately improving overall classification performance.




 


\section{Conclusion \& Future Work}
\label{sec:illust}
In this study, we have investigated the potential of image SR as a pre-processing step for improving multi-label scene classification in RS. 
Our findings reveal that across diverse SR architectures (e.g., SRResNet, HAT, SeeSR, RealESRGAN) and classification backbones (ResNet-50, ResNet-101, ResNet-152, Inception-v4), SR-based enhancements can yield notable gains in multiple evaluation metrics, including accuracy, Hamming Loss, One-Error, F1-Score, and Macro F2 Score.
Notably, SRResNet consistently boosted performance in shallower models (ResNet-50, ResNet-101), whereas the attention-based HAT approach aligned more effectively with deeper architectures (ResNet-152, Inception-v4).
In conclusion, this study bridges the gap between SR and multi-label classification in satellite imagery, offering a robust framework for improving remote sensing applications.

Future work should also analyze the effect of using SR models trained on satellite images for multi-label scene classification.

\subsection*{Acknowledgements}
This work was supported by the BMBF project SustainML (Grant 101070408) and by Carl Zeiss Foundation through the Sustainable Embedded AI project (P2021-02-009).



\clearpage
\bibliographystyle{IEEEbib}
\bibliography{refs}

\end{document}